\documentclass{CVM}

\def\Mymth{OPUS-V2}

\usepackage{xcolor}         % colors
\usepackage{makecell}
\usepackage{colortbl}
\usepackage{hyperref}
\usepackage{cleveref}
\usepackage{anyfontsize}
\usepackage{bbding}

\crefname{figure}{Fig.}{Figs.}
\crefname{table}{Tab.}{Tabs.}

\def\model{OPUS-V2}
\def\lmodule{point-voxel transformation}
\def\smodule{PVT}
\def\riou#1{$\text{rayIoU}_\text{#1}$}
\def\eg{\textit{e.g.}}

\def\etal{\textit{et al.}}

\CVMsetup{
type      = {Research/Review Article},
doi       = {CVM.XXXX},
title     = {\model{}: Bridging the Gap between Sparse Points and Dense Voxels},
author    = {Jiabao Wang$^{1}$, Qiang Meng$^{2}$, Liujiang Yan$^{3}$, Ke Wang$^{3}$, Qibin Hou$^{1}$\cor{}, Ming-Ming Cheng$^{1}$},
runauthor = {Jiabao Wang, Qiang Meng, Liujiang Yan, Ke Wang, Qibin Hou, Ming-Ming Cheng},
abstract  = {
The point-based occupancy prediction paradigm has achieved an attractive trade-off between accuracy and efficiency by modeling 3D space sparsely.
However, its predictions  inherently mismatch  the dense voxel-based occupancy required by self-driving systems, necessitating hand-crafted heuristics during training and inferencing that limit  final performance.
To overcome these limitations, we propose \model{}, a novel framework built upon the pioneering OPUS (occupancy prediction using a sparse set) point-based approach.
\model{} incorporates a lightweight \lmodule{}  module behind the decoder to adaptively map sparse predictions into the dense voxel space, eliminating the need for suboptimal operations and improving model accuracy.
Furthermore, our architecture decouples feature and occupancy generation processes, allowing \model{} to adapt to arbitrary occupancy resolutions. 
\model{} achieves a state-of-the-art rayIoU of 44.0 on the Occ3D dataset.
On the more challenging OpenOccupancy dataset, it attains a competitive 16.4 mIoU while running in real-time at 20.6 FPS.
Our code and models are publicly available at \url{https://github.com/NK-JittorCV/nk-occupancy}.
},
keywords  = {autonomous driving, occupancy prediction, sparse prediction, transformer},
copyright = {The Author(s)},
}

\begin{document}

\maketitle

\enlargethispage{-3pt}
\begin{figure}[b] \vskip -1mm
\small\renewcommand\arraystretch{1.3}
\begin{tabular}{p{80.5mm}} \toprule\\ \end{tabular}
\vskip -4.5mm \noindent \setlength{\tabcolsep}{1pt}
\begin{tabular}{p{3.5mm}p{80mm}}
$1\quad $ & VCIP, Nankai University, Tianjin, 300350, China. E-mail: Jiabao Wang, jbwang@mail.nankai.edu.cn; Qibin Hou, houqb@nankai.edu.cn; Ming-Ming Cheng, cmm@nankai.edu.cn.\\
$2\quad $ & Momenta, Beijing, 102200, China. E-mail: Qiang Meng, qiang.meng@momenta.ai.\\
$3\quad $ & KargoBot, Beijing, 102200, China. E-mail: Liujiang Yan, yanliujiang@kargobot.ai; Ke Wang, wangke@kargobot.ai.\\
&\hspace{-5mm} Manuscript received: 2025-01-01; accepted: 2025-01-01\vspace{-2mm}
\end{tabular} \vspace {-3mm}
\end{figure}

% -----------------------------------------------------------------------------------------------introducion
\section{Introduction}

\begin{figure}
    \centering
    \includegraphics[width=\linewidth]{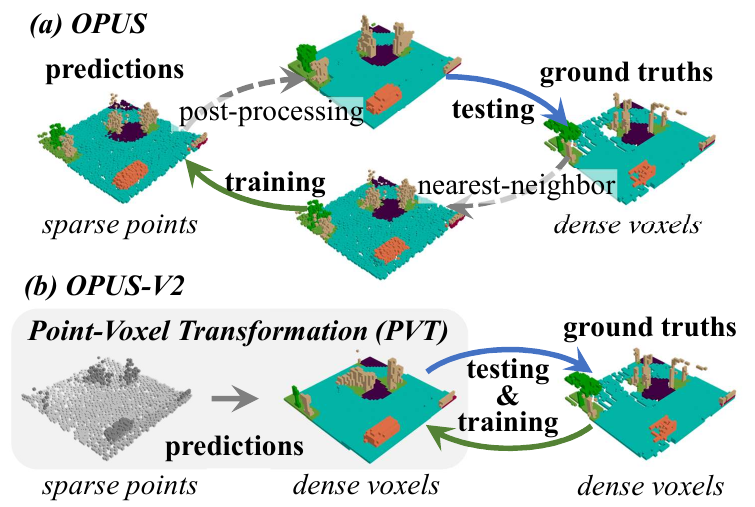}
    \caption{
        Comparison of  OPUS and \model{} pipelines. 
        (a) OPUS needs suboptimal operations, such as a nearest-neighbor strategy in training and handcrafted post-processing in testing.
        In contrast, (b) \model{} leverages \smodule{} to adaptively transform sparse predictions into the voxel space, facilitating direct interaction between predictions and ground truth.
    }\label{fig:gap}
\end{figure}

%%%RRM 
Predicting the occupancy status within a voxelized 3D environment~\cite{tian2024occ3d, wang2023openoccupancy} facilitates the recognition of unconventional obstacles, thereby significantly enhancing the safety of autonomous driving systems.
For a long time, this task has been treated as voxel-wise dense semantic segmentation~\cite{cao2022monoscene, li2023fb, zhang2023occformer, huang2023tri}, which is widely criticized for massively wasting  computational resources in unoccupied regions~\cite{liu2023fully}.
Recent advances have moved beyond the voxel-based paradigm to model the 3D world using sparse descriptors, including point sets~\cite{wang2024opus, dang2025sparseworld, li2025enhancing}, Gaussians~\cite{huang2024gau, huang2025gaussianformer, kerbl3Dgaussians}, and superquadrics~\cite{barr1981superquadrics}.
Among these representations, point-based modeling provides precise occupancy boundaries while maintaining high sparsity, enabling an effective trade-off between accuracy and efficiency.

\begin{figure*}[t]
    \centering
    \includegraphics[width=\linewidth]{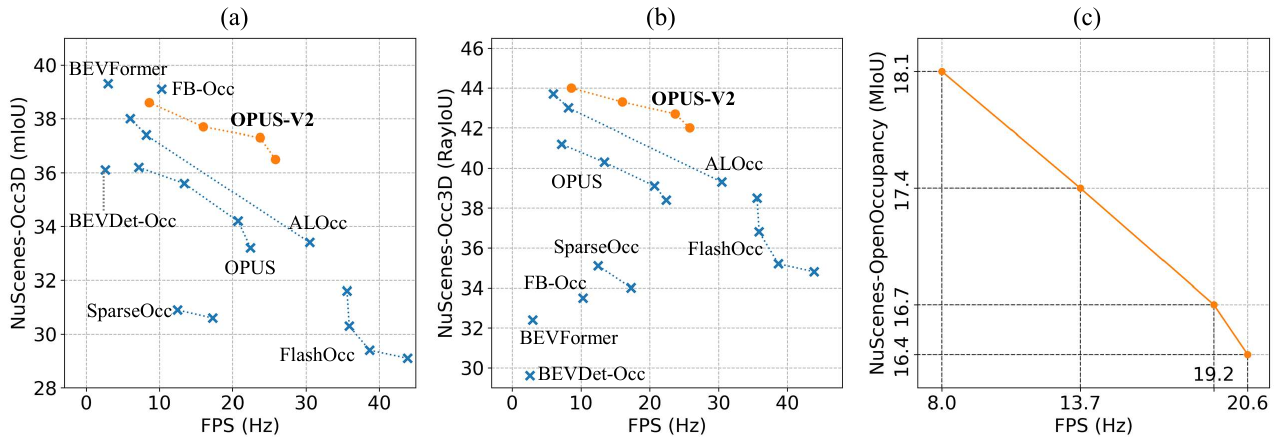}
    \caption{
        Comparison between \model{} and previous methods on the Occ3D and OpenOccupancy datasets, evaluated using both mIoU and rayIoU metrics. 
        FPS measurements were conducted on an A100 GPU with PyTorch FP32 precision, except for ALOcc~\cite{chen2025alocc}, for which FPS was measured on a 4090 GPU and is included for reference.
    }\label{fig:table}
\end{figure*}

However, autonomous driving systems ultimately require dense voxel-based predictions for downstream planning~\cite{elfes2002using, moravec1985high}, necessitating a conversion between point and voxel representations.
As illustrated in \cref{fig:gap}(a), current point-based approaches utilize deficient heuristics to bridge this representation gap, causing several problems.
(i) The one-to-one correspondence between the predictions and the ground truth no longer exists.
Models must employ a hand-crafted label assignment strategy during training, \eg nearest-neighbor matching, which typically leads to erroneous supervisory signals and renders the learning process vulnerable to noise.
2) During inferencing, transforming point predictions into the dense voxel grid requires additional manual post-processing, such as voxelization, dilation, and erosion.
These operations are inherently approximate and may degrade the fidelity of the reconstructed spatial occupancy.
As a result, direct interaction between predictions and annotation is obstructed by heuristic operations in both training and inferencing stages, implicitly limiting the attainable performance of the model.

To overcome these limitations, we propose \model{}, a new point-based approach which integrates the pioneering OPUS (occupancy prediction using a sparse set) with direct voxel-based occupancy generation, enabling seamless interaction with voxel-based ground truth.
Inspired by point cloud processing techniques~\cite{zhou2018voxelnet, yan2018second}, \model{} introduces a new \lmodule{} (\smodule{}) module to convert 3D points and their corresponding features into voxel-based occupancy in a learnable manner.
The \smodule{} module is lightweight, consisting only of two VFE layers~\cite{yan2018second} and two sparse convolution layers~\cite{graham2014spatially}, ensuring the overall efficiency of \model{}.
Despite its simplicity, the \smodule{} module provides two key advantages.
First, as illustrated in \cref{fig:gap}(b), the uniform representation of predictions and ground truth eliminates all manual operations required in the original OPUS, thus increasing the upper performance bound of the model.
Second, \model{} decouples feature extraction and occupancy generation into decoders and \smodule{}, respectively.
This design confines resolution scaling of occupancy predictions within \smodule{}, enabling \model{} to expand its perceptual range and granularity effortlessly.

We have conducted comprehensive experiments on the Occ3D~\cite{tian2024occ3d} and OpenOccupancy~\cite{wang2023openoccupancy} datasets.
As  Figs.~\ref{fig:table}(a) and \ref{fig:table}(b) show, on Occ3D, our lightest \model{} outperforms all original OPUS models in both accuracy and speed, underscoring the importance of the \smodule{} module. 
Our largest \model{} achieves a remarkable mIoU of 38.6 and the highest rayIoU of 44.0, setting a new state-of-the-art using Occ3D.
To further validate \model{}'s flexibility in spatial scaling, we evaluated it on OpenOccupancy, which features annotations with broader coverage and finer granularity. 
Despite generating significantly more voxels, \model{} maintained real-time performance at 20.6 FPS and achieved a leading mIoU of 16.4, as demonstrated in \cref{fig:table}(c), highlighting its efficiency in large-scale occupancy generation.

In summary, the contributions of this paper are as follows:
\begin{itemize}
    \item 
    The point-based paradigm requires manual alignment of sparse point-based predictions and dense voxel-based ground truth.
    In this paper, we propose \model{}, a novel architecture equipped with a lightweight \smodule{} module that seamlessly converts sparse points into voxel space, removing the need for manual operations while facilitating occupancy spatial scaling.

    \item Extensive experiments on the Occ3D-nuScenes dataset reveal that \smodule{}  significantly improves model performance.
    Our lightest variant surpasses all original OPUS models of every size. 
    Furthermore, \model{} achieves state-of-the-art results on the OpenOccupancy dataset with a real-time inference speed of 20.6 FPS, underscoring its exceptional efficiency and scalability.
\end{itemize}

%---------------------------------------------------------------------------related work
\section{Related Work}

\subsection{Voxel-based Occupancy Prediction}
The occupancy prediction task requires the model to perceive 3D spatial geometry and semantics based on voxels, which are intuitive for analysis and convenient for planning use.
Following Tesla AI Day 2022~\cite{tesla}, this task has recently attracted significant interest from both academic and industrial communities.
Many benchmarks~\cite{tian2024occ3d,wang2023openoccupancy,wei2023surroundocc,tong2023scene} have been proposed by voxelizing LiDAR points in standard self-driving datasets~\cite{caesar2020nuscenes,sun2020waymo,behley2019semantickitti,caesar2021nuplan}.
Meanwhile, a series of advances have achieved occupancy prediction by constructing dense 3D features for voxel grids.

However, this dense voxel-based paradigm suffers severely from high computational complexity.
To alleviate this problem, many works explore more efficient voxel representations.
In particular, FlashOcc~\cite{yu2023flashocc, yu2024panoptic} and TPVFormer~\cite{huang2023tri} compress the dense feature using the bird's-eye view (BEV) and tri-perspective view (TPV) representation for model efficiency.
COTR~\cite{ma2023cotr} down-samples the large 3D features and restores the original resolution in the final stage.
Recently, several works have leveraged transformers to refine complex geometric information.
For instance, VoxFormer~\cite{li2023voxformer} initiates sparse queries through a pre-task of depth estimation to focus on feature refinement at occupied regions.
SparseOcc proposed by Liu \etal~\cite{liu2023fully} uses a series of decoders to filter out empty cells and only predicts the occupation statuses of the retained voxels.
While these approaches have succeeded in reducing computational costs, they either sacrifice 3D features or require intricate space modeling, leading to an inferior trade-off between accuracy and efficiency.

\subsection{Sparse Representation of Occupancy}
Recently, many works have proposed alternative sparse occupancy representations to achieve greater efficiency.
Some of them describe occupancy as Gaussian kernels, inspired by the popular Gaussian splatting method~\cite{kerbl3Dgaussians}.
In particular, GaussianFormer~\cite{huang2024gau} first uses queries to aggregate image information via attention, then generates Gaussian kernels.
The subsequent GaussianFormer-2~\cite{huang2025gaussianformer} further reduces the number of kernels, thus improving its efficiency.
GaussianAD~\cite{zheng2024gaussianad} finally provided the first Gaussian-based end-to-end driving model.
In addition, the Gaussian representation is widely used in the weakly supervised and self-supervised occupancy prediction tasks, due to its ability to connect 3D and 2D spaces.
For example, GSRender~\cite{sun2024gsrender} generates Gaussian kernels in each voxel and weakly supervises them using 2D depth and semantic maps.
The Gaussian-based representation is more flexible than voxels, but its inductive ellipsoidal shape limits the modeling of diverse structures.

Meanwhile, OPUS~\cite{wang2024opus} formulates occupancy prediction as a point set prediction task, establishing a new point-based paradigm.
Sparse points can express arbitrary shapes, therefore eliminating the need for complex space modeling or sparsification procedures.
Following this paradigm, ODG~\cite{shi2025odg} combines point and Gaussian representations and further divided queries into dynamic and static components to process movable objects.
SparseWorld~\cite{dang2025sparseworld} builds a 4D occupancy world model by modeling occupancy as points.
DiScene~\cite{li2025enhancing} extends the point-based paradigm to the indoor occupancy prediction task.
Despite these successes, the point-based paradigm contains many manual operations, which limit their accuracy.

\subsection{Set Prediction with Transformers}
Our method is based on the point-based occupancy prediction paradigm OPUS, which inherits the end-to-end set prediction framework of SparseBEV~\cite{liu2023sparsebev}.
The set prediction framework was first introduced by DETR~\cite{carion2020end}, in which a collection of sparse queries produces an unordered set of outputs through feature interactions in transformer decoders.
Leveraging the strong modeling capacity of transformers, DETR dispenses with laborious post-processing steps such as NMS, achieving a clean end-to-end pipeline.  
Following DETR, many variants~\cite{zhu2020deformable,meng2021conditional,liu2022dab,li2022dn,wang2022anchor,zhang2022dino,sun2021rethinking} have been proposed to enhance performance and training efficiency.  
This sparse-query paradigm has also demonstrated its effectiveness in 3D object detection~\cite{wang2022detr3d,liu2022petr,lin2022sparse4d,liu2023sparsebev,wang2023exploring}, where each query represents the 3D attributes of a specific object.
However, transformer-based set prediction has been largely restricted to object detection scenarios, primarily due to constraints on number of queries.
OPUS effectively reduces the number of queries by predicting multiple points within a single query, marking the first introduction of set prediction into occupancy tasks.

%-----------------------------------------------------------------------Method--------
\section{Method}

\subsection{Preliminaries}
In this section, we first briefly revisit the point-based paradigm proposed by OPUS and then analyze the inherent problem of heuristic operations.

\subsubsection{Point-based occupancy prediction paradigm}
To reduce redundancy in voxel-based methods, OPUS  reformulates occupancy prediction as a point set prediction task.
Given voxel-based annotations $\mathbf{O}_\text{g} \in \mathbb{R}^{H \times W \times Z}$, the model first collects occupied voxels as a set of points $\mathcal{V}_\text{g}=\{(\mathbf{p}_\text{g}^n, \mathbf{c}_\text{g}^n)\}_{n=1}^{V_\text{g}}$.
In each entry  $(\mathbf{p}_\text{g}, \mathbf{c}_\text{g}) \in \mathcal{V}_\text{g}$, $\mathbf{p}_\text{g}$ represents the 3D coordinates of a voxel center, while $\mathbf{c}_\text{g}$ stores the semantic class of the voxel.
Consequently, the goal of OPUS is to generate $\mathcal{V}=\{(\mathbf{p}^n, \mathbf{c}^n)\}_{n=1}^{V}$ to mimic the distribution of $\mathcal{V}_\text{g}$.

In pursuit of training efficiency, OPUS decouples the task into separate regression and classification objectives.
Specifically, $\mathcal{V}$ and $\mathcal{V}_\text{g}$ are divided into the set of positions $\mathcal{P}=\{\mathbf{p}^n\}_{n=1}^V$ and semantic labels $\mathcal{C}=\{\mathbf{c}^n\}_{n=1}^V$, respectively.
For regression, OPUS employs the well-proven chamfer distance loss to minimize discrepancy, formulated as:
\begin{align}
    \text{CD}(\mathcal{P}, \mathcal{P}_\text{g}) =& \frac{1}{V}\sum\limits_{\mathbf{p} \in \mathcal{P}} D(\mathbf{p}, \mathcal{P}_\text{g}) + \frac{1}{V_\text{g}}\sum\limits_{\mathbf{p}_\text{g} \in \mathcal{P}_\text{g}} D(\mathbf{p}_\text{g}, \mathcal{P}), \nonumber \\
    &\text{ where } D(\mathbf{x}, \mathcal{Y}) = \min_{\mathbf{y} \in \mathcal{Y}}||\mathbf{x} - \mathbf{y}||_1.
\end{align}
Regarding the classification objective, OPUS leverages the nearest-neighbor strategy to construct a proxy $\hat{\mathcal{V}}_\text{g}$, where
\begin{equation}
    \hat{\mathcal{V}}_\text{g} = \left\{
        (\mathbf{p},\hat{\mathbf{c}}) \bigg| \mathbf{p} \in \mathcal{P}_\text{p}, \hspace{4pt}
        (\hat{\mathbf{p}},\hat{\mathbf{c}})=\mathop{\arg\min}\limits_{(\mathbf{p}_\text{g}, \mathbf{c}_\text{g})\in \mathcal{V}_\text{g}} \|\mathbf{p}_\textbf{g} - \mathbf{p}\|_2
    \right\}.
    \label{eq:label_cls}
\end{equation}
The $\hat{\mathcal{C}}_\text{g}=\{\hat{\mathbf{c}}^n\}_{n=1}^V$ are used to supervise the learning of $\mathcal{C}$.

During testing, OPUS applies a series of post-processing steps to convert sparse point-based predictions into voxel-based occupancy.
The model first removes low-quality points from $\mathcal{V}$ and then voxelizes the remainder into a dense occupancy $\hat{\mathbf{O}}\in \mathbb{R}^{H \times W \times Z}$.
A denser prediction is produced by performing 3D dilation and erosion through max-pooling operations, which can be described as
\begin{equation}
    \begin{split}
        \hat{\mathbf{O}}' &= \text{MaxPool3D}(\hat{\mathbf{O}}) \quad \text{(dilation)}, \\
        \hat{\mathbf{O}}' &= \mathbf{1} - \text{MaxPool3D}(\mathbf{1} - \hat{\mathbf{O}}') \quad \text{(erosion)}.      
    \end{split}
\end{equation}
Finally, the evaluation process compares the post-processed predictions $\hat{\mathbf{O}}$ with the ground truths $\mathbf{O}_\text{g}$.

\subsubsection{The problem of heuristic operations}
By reformulating occupancy prediction as a point set prediction task, OPUS avoids complex model design and significantly improves efficiency.
However, this task reformulation also leads to a gap between predictions and ground truths.
As stated above, OPUS predicts a set of sparse points $\mathcal{V}$, which cannot be directly associated with dense $\mathbf{O}_\text{g}$.
Therefore, the model must utilize hand-crafted operations, such as the nearest-neighbor strategy to construct the proxy $\hat{\mathcal{V}}_\text{g}$ in training and the post-processing for producing $\hat{\mathbf{O}}_\text{p}$, to align the format of predictions and ground truths, so that training and testing can proceed.

\begin{figure}
    \centering
    \includegraphics[width=\linewidth]{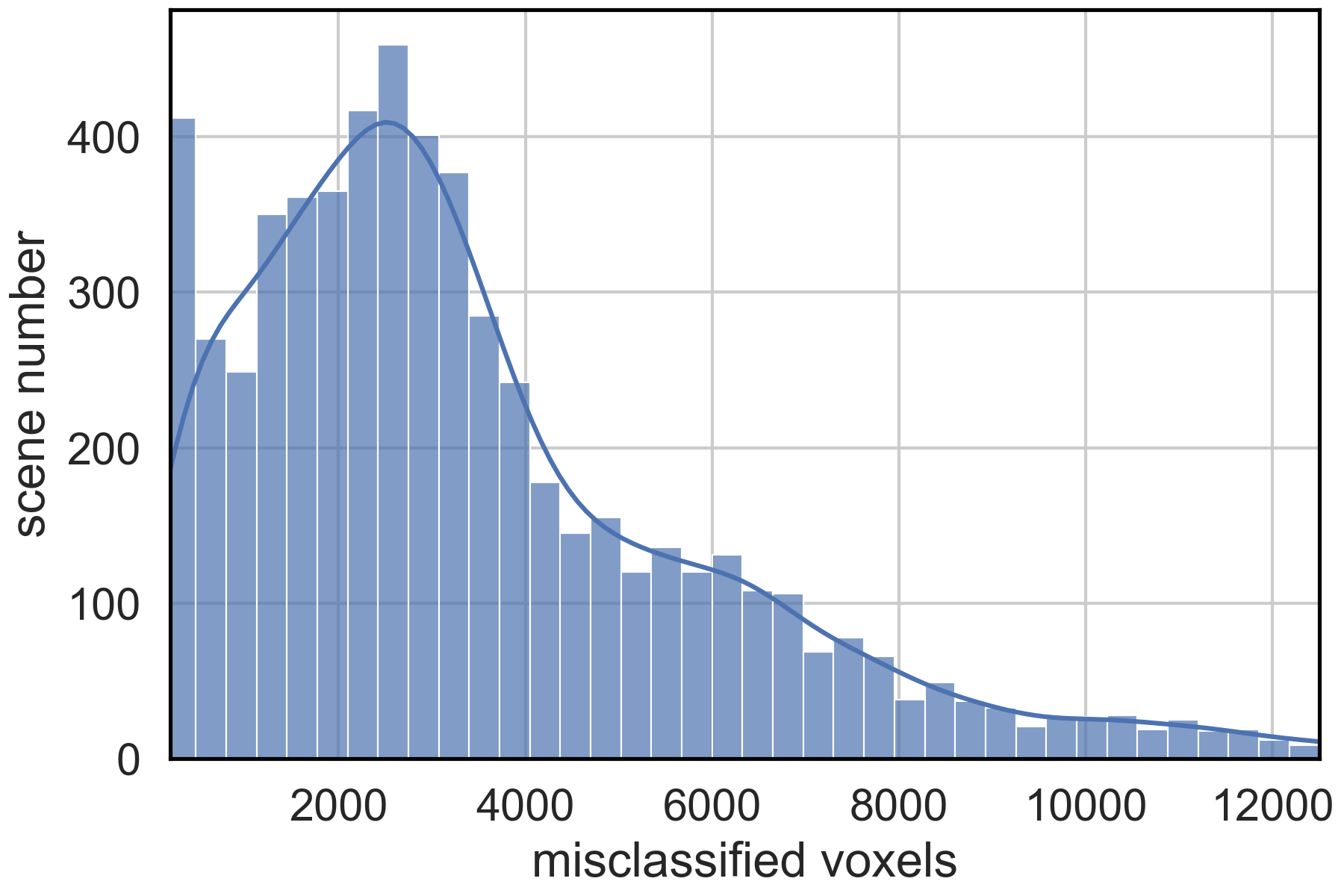}
    \caption{
        Numbers of mislabeled voxels in post-processing on the Occ3D dataset.
        We count mislabeled voxels whose occupancy status is correctly represented by predicted points but misclassified in post-processing. 
    }\label{fig:stat}
\end{figure}

However, these hand-made operations are inherently suboptimal:
the nearest-neighbor strategy only takes into account the label of the closest occupied voxel, and ignores the surrounding information.
Moreover, its local optimum property makes the proxy $\hat{\mathcal{V}}_\text{g}$ sensitive to prediction variants, leading to training instability.
The post-processing pipeline is also approximate and inaccurate.
We analyze the number of voxels whose occupancy status can be correctly represented by point predictions of OPUS-L but incorrectly stated in the final $\hat{\mathbf{O}}$in \cref{fig:stat}: the post-processing mislabels more than 2K voxels in most scenes and, in some cases, over 10K voxels, hindering OPUS from performing better.

\begin{figure*}[t]
    \centering
    \includegraphics[width=\textwidth]{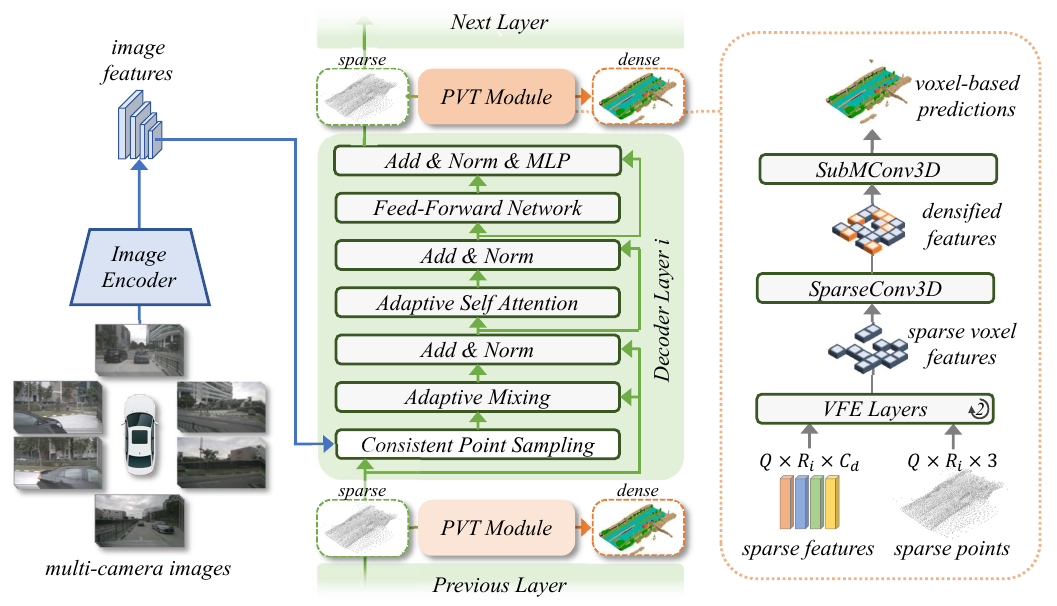}
    \caption{
        \model{} employs a three-stage pipeline:
        (1) an image encoder to extract multi-view image features, 
        (2) a series of decoders to produce 3D sparse points with corresponding latent features, and
        (3) a lightweight \smodule{} module that adaptively transforms these sparse features into dense voxel space to produce the final occupancy predictions.
    }\label{fig:structure}
\end{figure*}

\subsection{Design of \model{}}
To mitigate this problem, we introduce \model{}, which integrates the newly proposed \lmodule{} (\smodule{}) module to transform sparse point predictions into dense voxel-based occupancy.
As \cref{fig:structure} shows, \model{} has three main parts.
Initially, multi-view images are delivered to the image encoder to produce image features.
Subsequently, a set of queries $\mathcal{Q}$ and  corresponding positions $\mathcal{P}$ are initiated to iteratively aggregate geometry and semantic information from image features through a sequence of decoder layers.
Instead of predicting explicit classification scores in OPUS, each decoder in \model{} generates a set of points with latent features $\mathcal{F}$.
Finally, $\mathcal{P}$ and $\mathcal{F}$ are fed into \smodule{} to generate voxel-based occupancy predictions $\mathbf{O}$.
We next provide a detailed description of the decoder and \smodule{} in \model{}.

\subsubsection{Notation}
We denote the queries and point positions before feeding into decoders as $\mathcal{Q}_0, \mathcal{P}_0$, which are separately zero and uniformly initialized at the beginning of the training stage.
In the $i$-th decoder, query features and point positions are updated to $\mathcal{Q}_i, \mathcal{P}_i$, and meanwhile a set of latent features $\mathcal{F}_i$ is produced.
The length of these sets is $Q$, corresponding to the number of queries.
Each query feature $\mathbf{q}_i\in \mathcal{Q}_i, i \in \{0,  \cdots, 5\}$ is a vector of length 256.
Following OPUS, each $\mathbf{q}_i$ in \model{} predicts $R_i$ points.
Therefore, $\mathbf{p}_i\in \mathcal{P}_i$ and $\mathbf{f}_i \in \mathcal{F}_i$ have shapes of $R_i \times 3$ and $R_i \times F$, where $F$ represents the channel size of each point feature.

\subsubsection{Decoder structure}
The decoder in \model{} inherits its main structure from OPUS, with a minor upgrade to its consistent point sampling mechanism.
For a given query $\mathbf{q}_{i-1}\in \mathcal{Q}_{i-1}$ and  corresponding point predictions $\mathbf{p}_{i-1}\in \mathcal{P}_{i-1}$, the $i$-th decoder first extracts image features related to $\mathbf{q}_{i-1}$ by consistent point sampling.
Specifically, the model samples $S$ points using:
\begin{equation}
    \mathbf{s}_{i-1} = \mu(\mathbf{p}_{i-1}) + \mathbf{t} \cdot \sigma(\mathbf{p}_{i-1}) + \phi(\mathbf{q}_{i-1}),
\end{equation}
where $\mu(\cdot)$ and $\sigma(\cdot)$  calculate the mean and standard deviation of the input positions, and $\phi(\cdot)$ is a linear layer to generate sampling offsets from the query features.
Instead of directly multiplying $\phi(\mathbf{q}_{i-1})$ and $\sigma(\mathbf{p}_{i-1})$ as in OPUS, \model{} employs $S$ learnable prototype sampling points $\mathbf{t}$ re-weighted by $\sigma(\mathbf{p}_{i-1})$ to inherit the dispersion degree from $\mathbf{p}_{i-1}$, which prevents gradient vanishing in $\phi(\cdot)$.
These sampled points are finally projected into the 2D image space to interpolate the related features.
Subsequently, the query feature is updated to $\mathbf{q}_{i}$ with  adaptive mixing of the sampled features and queries, self-attention between all queries, and FFN, which are analogous to those in SparseBEV~\cite{liu2023sparsebev}.
In the end, two branches of MLP applied to $\mathbf{q}_i$ generate the offsets $\Delta \mathbf{p}_{i}$ (of size $R_i \times 3$) used to update point positions through $\mathbf{p}_i = \mu(\mathbf{p}_{i-1}) + \Delta \mathbf{p}_{i}$, and the latent features of each point $\mathbf{f}_{i}$ (size $R_i \times F$).

\begin{figure*}[t]
    \centering
    \includegraphics[width=\textwidth]{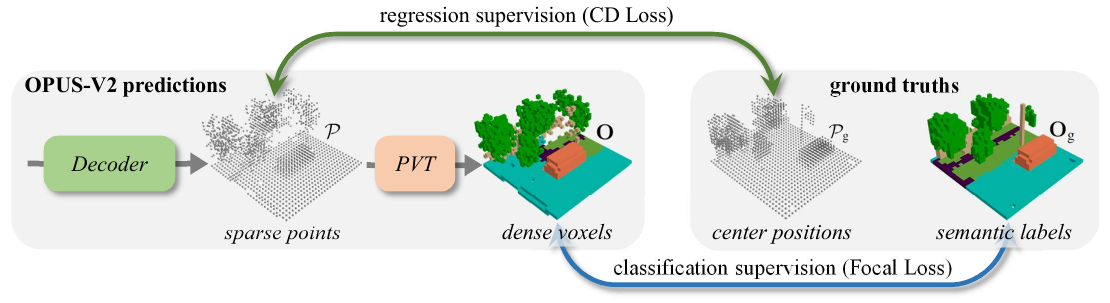}
    \caption{
        Training objectives in \model{}.
        We separately supervise the sparse and dense predictions of decoders and \smodule{} by the center points of occupied voxels and semantic labels of voxelized annotations.
    }\label{fig:loss}
\end{figure*}

\subsubsection{The \lmodule{} module}
The sparse points predicted by the decoder cannot be directly associated with the ground truth.
In \model{}, we place the \lmodule{} (\smodule{}) module after the decoder to adaptively transform sparse point predictions into voxel-based occupancy.
For given predictions $\mathbf{p}_i \in \mathcal{P}_i$ and $\mathbf{f}_i \in \mathcal{F}_i$ of the $i$-th decoder, we first construct point features $\mathbf{f}'_i \in \mathcal{F}'_i$ that contain position and semantic information. These can be formulated as:
\begin{equation}
    \mathbf{f}'_i = [\mathbf{p}_i-\mu(\mathbf{p}_i),\ \mathbf{p}_i-d(\mathbf{p}_i),\ \mathbf{f}_i].
\end{equation}
Here, we incorporate relative positions from the query center $\mu(\mathbf{p}_i)$ and the corresponding voxel center $d(\mathbf{p}_i)$ into the point features, which has been shown to be beneficial in experiments.
Subsequently, a series of dynamic VFE~\cite{yan2018second} layers aggregate the point features $\mathcal{F}'_i$ into voxel-wise representations.
For points within the same voxel, these layers first refine individual point features using a fully connected network (FCN) and then apply max-pooling to generate a single feature vector per voxel. 
The generated voxel features are organized as a sparse tensor.
Ultimately, to efficiently generate the final voxel-based predictions $\mathbf{O}_i$, we employ two sparse convolution layers~\cite{graham2014spatially} with a kernel size of 3.
The first layer is a SparseConv3D, which propagates information to empty voxels, analogous to the densification step in OPUS. 
This is followed by a SubMConv3D layer that performs computations only on non-empty voxels, thereby conserving computational resources.

The \smodule{} module serves as a bridge between the sparse point predictions of decoders and the dense voxel-based occupancy required for self-driving systems, which assists \model{} in two ways.
\begin{itemize}
    \item \emph{Omitting manual operations.}
    With the help of the \smodule{} module, \model{} can generate voxel-based occupancy, enabling direct interaction between predictions and ground truth rather than building handcrafted intermediates, thus improving  model accuracy.
    For example, \model{} no longer needs to construct a proxy $\hat{\mathcal{V}}_\text{g}$ for classification training.
    The sparse point predictions are voxelized and densified in a learnable manner, eliminating the need for hand-crafted post-processing.
    
    \item \emph{Flexibility in spatial scaling of occupancy.}
    In addition, \model{} decouples feature extraction and dense occupancy generation.
    The perceptual range and granularity of occupancy only affect the process of dense-occupancy generation in \smodule{}.
    In other words, while keeping the number of sparse points, \model{} can easily increase the occupancy resolution by modifying  hyperparameters in the lightweight \smodule{} module.
\end{itemize}

\subsubsection{Training objectives}
We simultaneously supervise the sparse and dense predictions of \model{} with the geometric and semantic   ground truth information as illustrated in \cref{fig:loss}.
For sparse point predictions, following the operation in OPUS, we implement a re-weighted chamfer distance loss between the sets of predicted points $\mathcal{P}$ and occupied voxel centers $\mathcal{P}_\text{g}$ so that the predictions have a similar geometric distribution to the ground truth, which can be described as:
\begin{align}
    &\text{CD}_\text{R}(\mathcal{P}, \mathcal{P}_\text{g}) = \frac{1}{V}\sum\limits_{\mathbf{p} \in \mathcal{P}} D_\text{R}(\mathbf{p}, \mathcal{P}_\text{g}) + \frac{1}{V_\text{g}}\sum\limits_{\mathbf{p}_\text{g} \in \mathcal{P}_\text{g}} D_\text{R}(\mathbf{p}_\text{g}, \mathcal{P}), \nonumber \\
    &\text{ where } D_\text{R}(\mathbf{x}, \mathcal{Y}) = W(d) d, \quad d=\min_{\mathbf{y} \in \mathcal{Y}}||\mathbf{x} - \mathbf{y}||_1.
\end{align}
Here, $W(d)$ is a re-weighting function penalizing points with large errors.
The specific design of $W(d)$ is provided later.
On the other hand, the dense voxel-based predictions $\mathbf{O}$ correspond one-to-one  to ground truth values $\mathbf{O}_\text{g}$.
Therefore, we directly employ a focal loss with class balance weights to optimize the learning of dense voxel-based predictions.
In the end, the training objectives of \model{} become:
\begin{align}
    L = \sum\limits_{i=1}^5(\text{CD}_R(&\mathcal{P}_i, \mathcal{P}_\text{g})+\text{FocalLoss}_R(\mathbf{O}_i, \mathbf{O}_\text{g})) \nonumber \\
    &+ \text{CD}_R(\mathcal{P}_0, \mathcal{P}_\text{g}),
\end{align}
which includes regression and classification losses at every stage, along with an additional $\text{CD}_R(\mathcal{P}_0, \mathcal{P}_\text{g})$ term to optimize the distribution of initial positions $\mathcal{P}_0$.

\subsubsection{\smodule{}  and other sparse voxel-based models}
The \smodule{} module, as a sparse voxel-based network, is structurally similar to previous sparse voxel-based occupancy predictors, like SparseOcc proposed by Tang \etal~\cite{tang2024sparseocc}.
However, their underlying approaches differ significantly.
Specifically, for occupancy prediction, \Mymth{} focuses on generating 3D points from 2D image features using transformer decoders.
With guidance from chamfer distance loss, these points can already accurately capture scene geometry and corresponding semantic information.
In contrast to heavy decoders,  \smodule{}  consists only of two VFE layers and two sparse convolutional layers.
Due to its limited capacity, it acts merely as a learnable post-processing step that refines the decoders' outputs into dense voxel-based predictions.
This stands in contrast to previous sparse voxel-based methods, which typically employ complex sparse voxel-based heads to extract per-voxel information from 3D dense features.
Their contribution lies in designing the head structure, rather than learning explicit geometry as \model{} does.

%---------------------------------------------------------------experiments-----------------------------
\section{Experiments}

\subsection{Experimental Setup}

\begin{table}[t]
    \caption{
        Configurations for different models.
        $Q$, $S$, $F$ represent the numbers of queries, sampling points in each query, and channel size of predicted point features.
        $R_i$ indicates the number of points predicted by a single query in the $i$-th decoder.
    }
    \label{tab:setting}
    \centering
    \setlength{\tabcolsep}{5pt}
    \begin{tabular}{l|ccc|cccccc}
    \toprule[0.9pt]
    \multirow{2}{*}{Model}  & \multirow{2}{*}{$Q$}  & \multirow{2}{*}{$S$} &\multirow{2}{*}{$F$}& \multicolumn{5}{c}{\#points in each stage}   \\
                            &                       &                      &                          & $R_1$ & $R_2$ & $R_3$ & $R_4$ & $R_5$  \\ \midrule
    \model{}-T       & 600                   & 4                    & 32                       & 8  & 16 & 32 & 64 & 128 \\
    \model{}-S       & 1200                  & 2                    & 32                       & 4  & 8  & 16 & 32 & 64 \\
    \model{}-M       & 2400                  & 2                    & 32                       & 2  & 4  & 8  & 16 & 32 \\
    \model{}-L       & 4800                  & 2                    & 32                       & 1  & 2  & 4  & 8  & 16 \\ \bottomrule[0.9pt]
    \end{tabular}
\end{table}

\subsubsection{Datasets}
We have evaluated our model on two large-scale driving benchmarks: Occ3D~\cite{tian2024occ3d} and OpenOccupancy~\cite{wang2023openoccupancy}.
Occ3D provides occupancy labels with a voxel size of $0.4$ m and a perceptual range of $\pm40$ m for 18 classes (1 free class and 17 semantic classes) on the nuScenes~\cite{caesar2020nuscenes} benchmark.
In contrast, OpenOccupancy provides much finer occupancy annotations for 17 classes (1 free class and 16 semantic classes), extending the perceptual range to $\pm51.4$ m and reducing the voxel size to $0.2$ m.
Following nuScenes, both datasets are split into 750, 150, and 150 driving scenes for training, validation, and testing, respectively.

\def\hl#1{\textbf{#1}}
\def\bevformer#1{BEVFormer#1~\cite{li2022bevformer}}
\def\renderocc#1{RenderOcc#1~\cite{pan2023renderocc}}
\def\bevdetocc#1{BEVDet-Occ#1~\cite{huang2021bevdet}}
\def\fbocc#1{FB-Occ#1~\cite{li2023fb}}
\def\sparseocc#1{SparseOcc#1~\cite{liu2023fully}}
\def\flashocc#1{P-FlashOcc#1~\cite{yu2024panoptic}}
\def\odg#1{ODG#1~\cite{shi2025odg}}
\def\stc#1{STCOcc#1~\cite{liao2025stcocc}}
\def\alocc#1{ALOcc#1~\cite{chen2025alocc}}
\def\daocc#1{DAOcc#1~\cite{yang2024daocc}}
\def\sgdocc#1{SGDOcc#1~\cite{duan2025sdgocc}}
\def\opus#1{OPUS#1~\cite{wang2024opus}}

\begin{table*}[t]
    \caption{
        Occupancy prediction performance on the Occ3D dataset. 
        The best result in each column is highlighted in {bold}.
        Models using temporal fusion over 8 and 16 frames are denoted by `8f' and `16f,' respectively.
        Some models implement CBGS~\cite{2019arXiv190809492Z}, which extends the training duration by nearly 4.5 times (marked with $^*$).
        All FPS results are sourced from their corresponding papers.
        Most were measured on an A100 GPU unless otherwise noted; ALOcc (marked with $^\dagger$) was measured on an RTX 4090.
     }
    \label{tab:occ3d}
    \centering
    \setlength{\tabcolsep}{2pt}
    \begin{tabular}{l|cccc|c|c|ccc|>{\columncolor{lightgray!20}}c|>{\columncolor{lightgray!20}}c}
    \toprule[0.9pt]
    Methods                & Input & Backbone & Image Size      & Epoch  & Vis. Mask    & mIoU      & \riou{1m} & \riou{2m} & \riou{4m} & \riou{}   & FPS \\ \midrule
    \renderocc{}           & C     & Swin-B   & $1408\times512$ & 12     & $\checkmark$ & 24.5      & 13.4      & 19.6      & 25.5      & 19.5      & - \\
    \bevformer{}           & C     & R101     & $1600\times900$ & 24     & $\checkmark$ & 39.3      & 26.1      & 32.9      & 38.0      & 32.4      & 3.0 \\
    \bevdetocc{}           & C     & R50      & $704\times256$  & 90     & $\checkmark$ & 36.1      & 23.6      & 30.0      & 35.1      & 29.6      & 2.6 \\
    \bevdetocc{ (8f)}      & C     & R50      & $704\times384$  & 90     & $\checkmark$ & 39.3      & 26.6      & 33.1      & 38.2      & 32.6      & 0.8 \\
    \fbocc{ (16f)}         & C     & R50      & $704\times256$  & 80$^*$ & $\checkmark$ & 39.1      & 26.7      & 34.1      & 39.7      & 33.5      & 10.3 \\
    \sparseocc{ (8f)}      & C     & R50      & $704\times256$  & 12     & -            & 30.9      & 28.0      & 34.7      & 39.4      & 34.0      & 17.3 \\
    \sparseocc{ (8f)}      & C     & R50      & $704\times256$  & 60     & -            & -         & -         & -         & -         & 37.7      & 17.3 \\
    \sparseocc{ (16f)}     & C     & R50      & $704\times256$  & 12     & -            & 30.6      & 29.1      & 35.8      & 40.3      & 35.1      & 12.5 \\
    \flashocc{-Tiny (1f)}  & C     & R50      & $704\times256$  & 24     & -            & 29.1      & 29.1      & 35.7      & 39.7      & 34.8      & \hl{43.9} \\
    \flashocc{ (1f)}       & C     & R50      & $704\times256$  & 24     & -            & 29.4      & 29.4      & 36.0      & 40.1      & 35.2      & 38.7 \\
    \flashocc{ (2f)}       & C     & R50      & $704\times256$  & 24     & -            & 30.3      & 31.2      & 37.6      & 41.5      & 36.8      & 35.9 \\
    \flashocc{ (8f)}       & C     & R50      & $704\times256$  & 24     & -            & 31.6      & 32.8      & 39.3      & 43.4      & 38.5      & 35.6 \\
    \odg{-T (8f)}          & C     & R50      & $704\times256$  & 100    & -            & 35.5      & -         & -         & -         & 39.2      & 20.1 \\
    \odg{-L (8f)}          & C     & R50      & $704\times256$  & 100    & -            & 38.2      & -         & -         & -         & 42.3      & 4.9 \\
    \stc{ (8f)}            & C     & R50      & $704\times256$  & 36     & -            & -         & 36.2      & 42.7      & 46.4      & 41.7      & -  \\
    \alocc{-2D-mini (16f)} & C     & R50      & $704\times256$  & 54$^*$ & -            & 33.4      & 32.9      & 40.1      & 44.8      & 39.3      & 30.5$^\dagger$  \\ 
    \alocc{-2D (16f)}      & C     & R50      & $704\times256$  & 54$^*$ & -            & 37.4      & 37.1      & 43.8      & 48.2      & 43.0      & 8.2$^\dagger$  \\
    \alocc{-3D (16f)}      & C     & R50      & $704\times256$  & 54$^*$ & -            & 38.0      & 37.8      & 44.7      & 48.8      & 43.7      & 6.0$^\dagger$  \\
    \daocc{}               & C\&L  & R50      & $704\times256$  & 27$^*$ & -            & -         & -         & -         & -         & 48.4      & 7.8$^\dagger$  \\
    \sgdocc{}              & C\&L  & R50      & $704\times256$  & 36     & $\checkmark$ & \hl{51.7} & -         & -         & -         & -         & -  \\ \midrule
    \opus{-T (8f)}         & C     & R50      & $704\times256$  & 100    & -            & 33.2      & 31.7      & 39.2      & 44.3      & 38.4      & 22.4 \\
    \opus{-S (8f)}         & C     & R50      & $704\times256$  & 100    & -            & 34.2      & 32.6      & 39.9      & 44.7      & 39.1      & 20.7 \\
    \opus{-M (8f)}         & C     & R50      & $704\times256$  & 100    & -            & 35.6      & 33.7      & 41.1      & 46.0      & 40.3      & 13.4 \\
    \opus{-L (8f)}         & C     & R50      & $704\times256$  & 100    & -            & 36.2      & 34.7      & 42.1      & 46.7      & 41.2      & 7.2 \\ \midrule
    \model{}-T (8f)        & C     & R50      & $704\times256$  & 50     & -            & 35.6      & 34.6      & 41.7      & 46.5      & 41.0      & 25.8 \\
    \model{}-S (8f)        & C     & R50      & $704\times256$  & 50     & -            & 36.6      & 35.7      & 42.9      & 47.3      & 42.0      & 23.7 \\
    \model{}-M (8f)        & C     & R50      & $704\times256$  & 50     & -            & 37.2      & 36.3      & 43.4      & 47.8      & 42.5      & 16.0 \\
    \model{}-L (8f)        & C     & R50      & $704\times256$  & 50     & -            & 38.4      & 37.5      & 44.4      & 48.7      & 43.6      & 8.6 \\ \midrule
    \model{}-T (8f)        & C     & R50      & $704\times256$  & 100    & -            & 36.5      & 35.8      & 42.8      & 47.3      & 42.0      & 25.8 \\
    \model{}-S (8f)        & C     & R50      & $704\times256$  & 100    & -            & 37.3      & 36.7      & 43.5      & 47.8      & 42.7      & 23.7 \\
    \model{}-M (8f)        & C     & R50      & $704\times256$  & 100    & -            & 37.7      & 37.2      & 44.3      & 48.5      & 43.3      & 16.0 \\
    \model{}-L (8f)        & C     & R50      & $704\times256$  & 100    & -            & 38.6      & \hl{38.0} & \hl{45.0} & \hl{49.2} & \hl{44.0} & 8.6 \\
    \bottomrule[0.9pt]
    \end{tabular}
\end{table*}

\def\mono#1{MonoScene#1~\cite{cao2022monoscene}}
\def\tpv#1{TPVFormer#1~\cite{huang2023tri}}
\def\sket#1{3DSketch#1~\cite{chen20203d}}
\def\aic#1{AICNet#1~\cite{li2020anisotropic}}
\def\lmsc#1{LMSCNet#1~\cite{roldao2020lmscnet}}
\def\jsc#1{LS3C-Net#1~\cite{yan2021sparse}}
\def\cconet#1{C-CONet#1~\cite{wang2023openoccupancy}}
\def\lconet#1{L-CONet#1~\cite{wang2023openoccupancy}}
\def\mconet#1{M-CONet#1~\cite{wang2023openoccupancy}}
\def\spocc#1{SparseOcc#1~\cite{tang2024sparseocc}}

\begin{table*}
    \setlength{\tabcolsep}{2.5pt}
	\centering
    \caption{
        Occupancy prediction performance on the OpenOccupancy dataset. 
        The best result in each column is highlighted in {bold}.
        The results of previous methods are taken from the papers of OpenOccupancy and SparseOcc proposed by Tang \etal
        FPS results of \model{} are measured on an A100 GPU.
    }\label{tab:openocc}
	\begin{tabular}{l|c| c c | c c c c c c c c c c c c c c c c | c }
	\toprule[0.9pt]
	Method
	& Input
	& IoU
    & mIoU
	& \rotatebox{90}{barrier} 
	& \rotatebox{90}{bicycle}
	& \rotatebox{90}{bus} 
	& \rotatebox{90}{car} 
	& \rotatebox{90}{const. veh.} 
	& \rotatebox{90}{motorcycle} 
	& \rotatebox{90}{pedestrian} 
	& \rotatebox{90}{traffic cone} 
	& \rotatebox{90}{trailer} 
	& \rotatebox{90}{truck} 
	& \rotatebox{90}{drive. suf.} 
	& \rotatebox{90}{other flat} 
	& \rotatebox{90}{sidewalk} 
	& \rotatebox{90}{terrain} 
	& \rotatebox{90}{manmade} 
	& \rotatebox{90}{vegetation}
    & FPS \\ \midrule
    \mono{}           & C    & 18.4      & 6.9       & 7.1       & 3.9       & 9.3       & 7.2       & 5.6       & 3.0       & 5.9       & 4.4       & 4.9       & 4.2       & 14.9      & 6.3       & 7.9       & 7.4       & 10.0      & 7.6       & -    \\
    \tpv{}            & C    & 15.3      & 7.8       & 9.3       & 4.1       & 11.3      & 10.1      & 5.2       & 4.3       & 5.9       & 5.3       & 6.8       & 6.5       & 13.6      & 9.0       & 8.3       & 8.0       & 9.2       & 8.2       & -    \\
    \cconet{}         & C    & 20.1      & 12.8      & 13.2      & 8.1       & 15.4      & 17.2      & 6.3       & 11.2      & 10.0      & 8.3       & 4.7       & 12.1      & 31.4      & 18.8      & 18.7      & 16.3      & 4.8       & 8.2       & -    \\
    \spocc{}          & C    & 21.8      & 14.1      & 16.1      & 9.3       & 15.1      & 18.6      & 7.3       & 9.4       & \hl{11.2} & 9.4       & 7.2       & 13.0      & 31.8      & 21.7      & 20.7      & 18.8      & 6.1       & 10.6      & -    \\ 
    \sket{}           & C\&D & 25.6      & 10.7      & 12.0      & 5.1       & 10.7      & 12.4      & 6.5       & 4.0       & 5.0       & 6.3       & 8.0       & 7.2       & 21.8      & 14.8      & 13.0      & 11.8      & 12.0      & 21.2      & -    \\
    \aic{}            & C\&D & 23.8      & 10.6      & 11.5      & 4.0       & 11.8      & 12.3      & 5.1       & 3.8       & 6.2       & 6.0       & 8.2       & 7.5       & 24.1      & 13.0      & 12.8      & 11.5      & 11.6      & 20.2      & -    \\
    \lmsc{}           & L    & 27.3      & 11.5      & 12.4      & 4.2       & 12.8      & 12.1      & 6.2       & 4.7       & 6.2       & 6.3       & 8.8       & 7.2       & 24.2      & 12.3      & 16.6      & 14.1      & 13.9      & 22.2      & -    \\
    \jsc{}            & L    & 30.2      & 12.5      & 14.2      & 3.4       & 13.6      & 12.0      & 7.2       & 4.3       & 7.3       & 6.8       & 9.2       & 9.1       & 27.9      & 15.3      & 14.9      & 16.2      & 14.0      & \hl{24.9} & -    \\
    \lconet{}         & L    & \hl{30.9} & 15.8      & 17.5      & 5.2       & 13.3      & 18.1      & 7.8       & 5.4       & 9.6       & 5.6       & \hl{13.2} & 13.6      & 34.9      & 21.5      & 22.4      & 21.7      & \hl{19.2} & 23.5      & -    \\\midrule
    \model{}-T        & C    & 27.4      & 16.4      & 17.5      & 7.5       & 16.2      & 18.9      & 10.4      & 11.7      & 6.9       & 6.5       & 8.2       & 14.9      & 39.3      & 27.3      & 25.4      & 23.2      & 11.7      & 16.4      & \hl{20.6} \\
    \model{}-S        & C    & 27.6      & 16.7      & 17.0      & 9.1       & 15.8      & 19.2      & 10.1      & 12.7      & 8.1       & 8.0       & 8.1       & 14.9      & 39.6      & 27.0      & 25.8      & 23.8      & 11.6      & 16.6      & 19.2 \\
    \model{}-M        & C    & 27.9      & 17.4      & 18.5      & 11.2      & 15.9      & 19.5      & 10.5      & 13.8      & 9.4       & 9.3       & 8.2       & 15.3      & 39.8      & 27.5      & 26.2      & 23.8      & 12.4      & 17.1      & 13.7 \\
    \model{}-L        & C    & 28.7      & \hl{18.1} & \hl{19.3} & \hl{11.5} & \hl{16.4} & \hl{19.9} & \hl{11.8} & \hl{15.0} & 10.0      & \hl{10.6} & 8.4       & \hl{15.8} & \hl{39.9} & \hl{27.7} & \hl{26.8} & \hl{24.5} & 13.9      & 18.0      & 8.0  \\
	\bottomrule[0.9pt]
	\end{tabular}
\end{table*}

\subsubsection{Metrics}
The commonly used mIoU metric was utilized for evaluation on both datasets.
In particular, Occ3D ignores unseen regions through camera masks, while OpenOccupancy removes areas labeled as `noise'.
Recently, a new \riou{}~\cite{liu2023fully} has been proposed as a remedy, as the mIoU metric can be easily overestimated.
Therefore, we also report \riou{} results for different distance thresholds (1 m, 2 m, and 4 m) on the Occ3D dataset, denoted  \riou{1m}, \riou{2m}, and \riou{4m}, respectively. 
The final \riou{} score is the average of these three values.

\subsubsection{Implementation details}
Following the common setting of the previous work~\cite{liu2023fully}, we resized images to $704 \times 256$ and extracted features using a ResNet50~\cite{he2016deep} backbone.
In addition to resizing and flipping image augmentations, a random occupancy flipping was implemented to enhance the diversity of the ground truth.
Like OPUS, we set up a series of models with 0.6K, 1.2K, 2.4K, and 4.8K queries (see \cref{tab:setting}), denoted  \model{}-T, \model{}-S, \model{}-M, and \model{}-L, respectively.
Queries at the same stage of one model predict an equal number of points.
The predicted number of points increases gradually across stages, totaling 76.8K points in the final stage.
Except for \model{}-T, all models sampled 2 points in each query in consistent point sampling.
All models were trained on 8 Nvidia A100-40G GPUs with a batch size of 8 using the AdamW~\cite{loshchilov2018decoupled} optimizer.
The learning rate increased to $1\times 10^{-4}$ over the first 500 iterations and then decayed using a cosine annealing~\cite{loshchilov2016sgdr} scheme. 
Unless otherwise stated, we trained all models for 50 epochs.

\subsubsection{Loss reweighting}
We adopted two reweighting strategies for the chamfer distance loss.
First, if a ground truth voxel was not occupied by any predicted point, we increased its weight by 5 to ensure that all ground truth voxels were covered by predicted points.
Then, for voxels belonging to categories with very low appearance frequency, such as `others', `bicycle', `construction vehicle', and `traffic cone', we also increase their weight by 5.
For the focal Loss, we assigned different weights to voxels based on category frequency.
Voxels from the most frequent categories, \eg, `drivable surface', `sidewalk', `manmade', and `terrain' received a weight of 1.
Categories with relatively low frequency, including `barrier', `bus', `car', `pedestrian', `trailer', `truck', and `other flat', were assigned a weight of 5.
The remaining infrequent categories, including `others', `bicycle', `construction vehicle', `motorcycle', and `traffic cone', were given a weight of 10.

\subsection{Main Results}

\subsubsection{Results on the Occ3D dataset}
We first compared the newly proposed \model{} to the original OPUS on the Occ3D-nuScenes dataset.
As reported in \cref{tab:occ3d}, \model{} achieved much better results than the original OPUS, where even the lightest \model{}-T  attained an mIoU of 36.5 and a rayIoU of 42.0, outperforming the heaviest original OPUS model.
The heaviest \model{}-L model achieved an mIoU of 38.6 and a rayIoU of 44.0, exceeding the performance of OPUS-L by 2.4 in mIoU and 2.8 in rayIoU.
In addition, due to the reduction of the decoders (5 in \model{}, 6 in OPUS) as reported in \cref{tab:setting}, the inferencing speed of \model{}  is also greater.
Currently, \model{}-T can reach 25.8 FPS, faster than OPUS-T by a large margin, and even \model{}-L with 4.8K queries can  run at 8.6 FPS.

A known limitation of the original OPUS is its slow convergence.
Our experiments indicated that this issue was significantly mitigated in \model{}, which achieved strong performance after only 50 epochs of training. 
As  \cref{tab:occ3d} shows, a 50-epoch \model{} achieved a rayIoU of 43.6, which is only 0.4 lower than its 100-epoch counterpart.
We attribute this improved training efficiency to the removal of the unstable nearest-neighbor matching strategy.

We further compared \Mymth{} to prior state-of-the-art methods. 
Our method not only achieved superior performance in terms of \riou{} at real-time speed, but also achieved competitive results in terms of mIoU.
Specifically, the \model{}-T variant achieved 25.8 FPS, outperforming most methods in speed except for the speed-optimized FlashOcc series, while surpassing FlashOcc in accuracy by up to 3.5 rayIoU. 
As the number of queries increased, \model{} outperformed all previous camera-only models while maintaining a commendable speed.
The largest variant, \model{}-L, achieved a state-of-the-art rayIoU of 44.0, surpassing the best prior model, ALOcc~\cite{chen2025alocc}, in both accuracy and speed. 
Notably, although trained without camera masks, \model{} achieves an mIoU of 38.6, a significant improvement over the original OPUS and only marginally lower (by less than 1 mIoU) than models that utilize  masks.

\subsubsection{Results on the OpenOccupancy dataset}
Compared to Occ3D, OpenOccupancy features a wider perceptual range and finer voxel granularity, presenting a significant challenge to model efficiency. 
We next tested \model{} on the OpenOccupancy dataset to further prove its flexibility in spatial scaling. 
As  \cref{tab:openocc} shows, the lightest \model{}-T  attained an mIoU of 16.4, even surpassing some previous LiDAR-based methods, including AICNet~\cite{li2020anisotropic}, LMSCNet~\cite{roldao2020lmscnet}, and L-CONet~\cite{wang2023openoccupancy}.
The heaviest model  reached 18.1 mIoU, establishing a new maximum for camera-based methods with a 4 mIoU advantage.

\model{} divides the processes of feature extraction and occupancy prediction into decoders and \smodule{} modules, allowing the model to increase the resolution of predicted occupancy at low cost.
In experiments, we transferred \model{} models from Occ3D to OpenOccupancy only by enlarging the \smodule{} module.
Despite the volume of occupancy predictions increasing by more than 8 times, \model{} still maintained real-time performance, with \model{}-T reaching 20.6 FPS and \model{}-L reaching 8.0 FPS, further demonstrating the high efficiency of \model{}.

\subsection{Ablation and Related Studies}
We next present some ablation and related studies of \model{}.

\begin{table}[t]
    \caption{
        Comparison of the sparse voxel-based method for different input image resolutions.
        The results of SparseOcc  by Tang \etal were obtained from their original paper.
        FPS results were measured on a RTX 4090 GPU.
    }
    \label{tab:comparison_standard}
    \centering
    \setlength{\tabcolsep}{4pt}
    \begin{tabular}{c|c|cc|cc}
    \toprule[0.9pt]
    models                      & Image Size      & IoU  & mIoU & Mem. (GB) & FPS \\ \midrule
    \multirow{2}{*}{\spocc{}}   & $704\times256$  & 21.8 & 14.1 & 13.0     & - \\
                                & $1600\times900$ & 20.4 & 14.6 & -        & - \\
    \multirow{2}{*}{\Mymth{-T}} & $704\times256$  & 24.8 & 15.1 & 2.3      & 24.2 \\
                                & $1600\times900$ & 26.2 & 15.7 & 11.6     & 6.9 \\ \bottomrule[0.9pt]
    \end{tabular}
    \label{rrm}
\end{table}

\subsubsection{Comparison of sparse voxel-based methods}
We compared the results of \Mymth{-T} and SparseOcc proposed by Tang \etal on the OpenOccupancy-nuScenes dataset under different input image resolutions. 
To ensure fairness, our model was also trained for 30 epochs using single-frame input.
Our experimental results are shown in  Table~\ref{rrm}.
With the same training strategy, \Mymth{-T} achieved 15.1 mIoU, outperforming SparseOcc, demonstrating the advantage of OPUS-V2 over standard sparse voxel-based methods.
As we increased the resolution of the input image, the performance of \Mymth{-T} further improved to 15.7, indicating that OPUS-V2-T benefits from higher-resolution inputs.
Meanwhile, even with the large input size of $1900\times 600$, \Mymth{-T} maintained 6.9 FPS, further demonstrating its efficiency.

\begin{table}[t]
    \caption{
        Ablation studies for different training and testing strategies.
        `NN' means using the proxy constructed by the nearest-neighbor in classification training.
        `\smodule{}' means employing the \smodule{} module in model.
    }
    \label{tab:strategy}
    \centering
    \setlength{\tabcolsep}{3.5pt}
    \begin{tabular}{c|cc|ccc|c}
    \toprule[0.9pt]
    index & NN         & \smodule{} & \riou{1m} & \riou{2m} & \riou{4m} & \riou{} \\ \midrule
    (I)   & \checkmark &            & 30.3      & 38.0      & 43.1      & 37.2 \\
    (II)  & \checkmark & \checkmark & 34.0      & 41.3      & 46.2      & 40.6 \\
    (III) &            & \checkmark & 34.6      & 41.7      & 46.5      & 41.0 \\ \bottomrule[0.9pt]
    \end{tabular}
\end{table}

\begin{table}[t]
    \caption{
        Ablation studies for components in point features.
        $\mathbf{f}$ denotes the semantic features predicted by decoders;
        $\mathbf{p}_\text{c}$ and $\mathbf{p}_\text{v}$ denote relative positions from  corresponding query centers and voxel centers, respectively.
    }
    \label{tab:features}
    \centering
    \setlength{\tabcolsep}{5pt}
    \begin{tabular}{ccc|ccc|c}
    \toprule[0.9pt]
    $\mathbf{f}$ & $\mathbf{p}_\text{c}$ & $\mathbf{p}_\text{v}$ & \riou{1m} & \riou{2m} & \riou{4m} & \riou{} \\ \midrule
    \checkmark   &                       &                       & 33.6      & 41.0      & 45.9      & 40.2 \\
    \checkmark   & \checkmark            &                       & 34.5      & 41.6      & 46.5      & 40.9 \\ 
    \checkmark   & \checkmark            & \checkmark            & 34.6      & 41.7      & 46.5      & 41.0 \\ \bottomrule[0.9pt]
    \end{tabular}
\end{table}

\subsubsection{Influence of manual operations}
We established three models with different training and testing on Occ3D to explore the influence of manual operations.
Model (I), like the original OPUS, only generated sparse point predictions and required manual operations in both training and testing.
In contrast, models (II) and (III) employed \smodule{} to directly generate voxel-based occupancy, eliminating the need for hand-crafted post-processing during testing.
As  \cref{tab:strategy} shows, the accuracies of models (II) and (III) are much better than that of model (I), demonstrating the effectiveness of replacing hand-crafted post-processing with the learning-based \smodule{}.
In addition, compared to model (II), model (III) further abandoned the proxy built by the nearest-neighbor strategy, leading to a slight improvement in the final result.
These results demonstrate that manual operations during training also hinder model performance.

\subsubsection{Influence of components in point features}
\cref{tab:features} lists the results of models with point features composed of different elements.
Using only semantic features yielded a rayIoU of 40.2, while incorporating both semantic and positional features improved the score to 41.0, underscoring the benefit of multi-modal information. 
Further adding both the relative positions from query centers and voxel centers provided an additional gain of 0.1 rayIoU. 
Based on these findings, our final configuration integrates all semantic and positional information to construct the point features.

\subsubsection{Effect of the number of decoders}
\begin{figure}[t!]
    \centering
    \includegraphics[width=\linewidth]{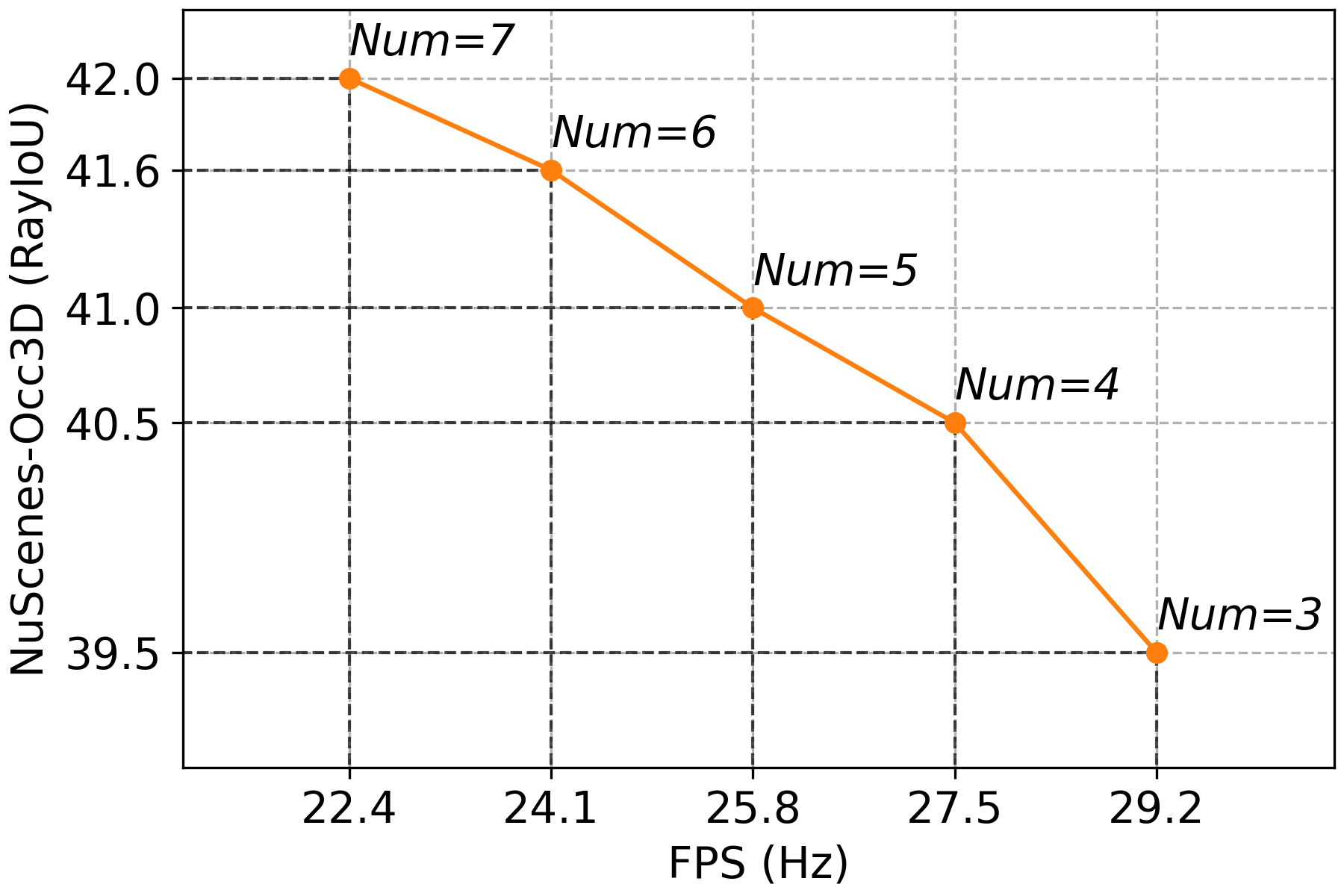}
    \caption{
        Varying the number of decoders.
        All models were trained for 50 epochs.
        FPS results were measured on an A100 GPU.
    }\label{fig:stage}
\end{figure}

We conducted a study on the Occ3D dataset using \model{}-T with different numbers of decoder layers, while keeping the final output point count constant for a fair comparison. 
As  \cref{fig:stage} shows, increasing the number of decoders from 3 to 7 leads to a consistent improvement in accuracy at the cost of reduced inferencing speed. 
This trend indicates that \model{}'s performance has not yet saturated, suggesting potential for further gains.
However, we found that a 5-decoder \model{} already surpasses all 6-decoder variants of the original OPUS. 
We therefore adopted this as the default configuration, achieving an optimal balance between speed and accuracy.

\subsubsection{Interaction between and sensitivity of different losses}
\begin{figure}[t!]
    \centering
    \includegraphics[width=\linewidth]{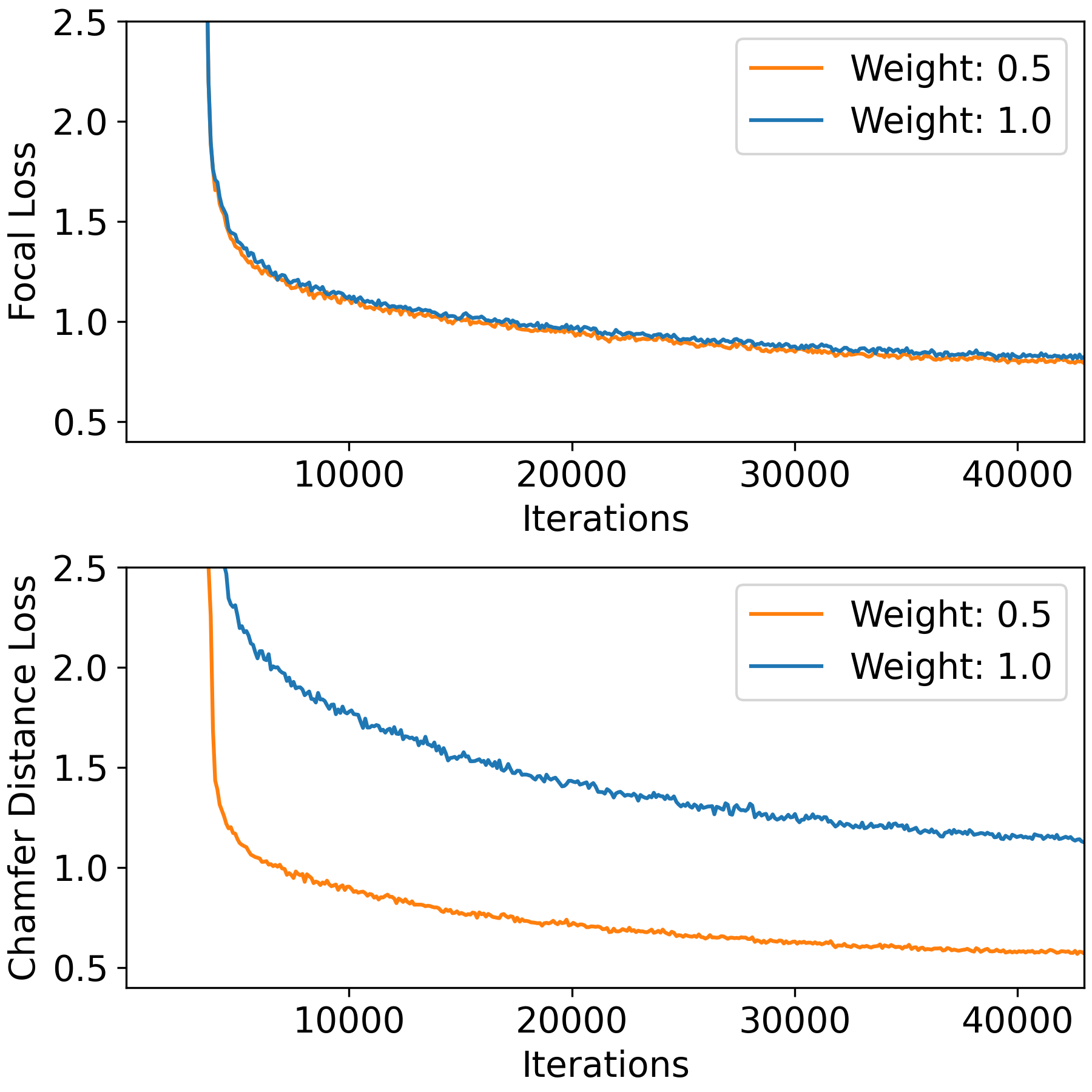}
    \caption{
        Convergence  of focal loss and chamfer distance loss with weights of chamfer distance loss set to 0.5 and 1.0.
    }\label{fig:loss_curve}
\end{figure}

To analyze the interaction between and sensitivity of the different loss components of \Mymth{}, we plotted  curves of  chamfer distance loss applied to the sparse predictions and focal loss applied to the dense predictions from the final decoder layer, for varying chamfer distance loss weights.
As  \cref{fig:loss_curve} shows, both losses decreased smoothly as  training  progressed.
To investigate their mutual influence, we doubled the weight of the chamfer distance loss.
In response, the focal loss curve exhibited only a modest increase.
This suggests a certain degree of competition between those losses, yet neither exerts a significant adverse impact on the other.

\begin{figure*}[t]
    \centering
    \includegraphics[width=\textwidth]{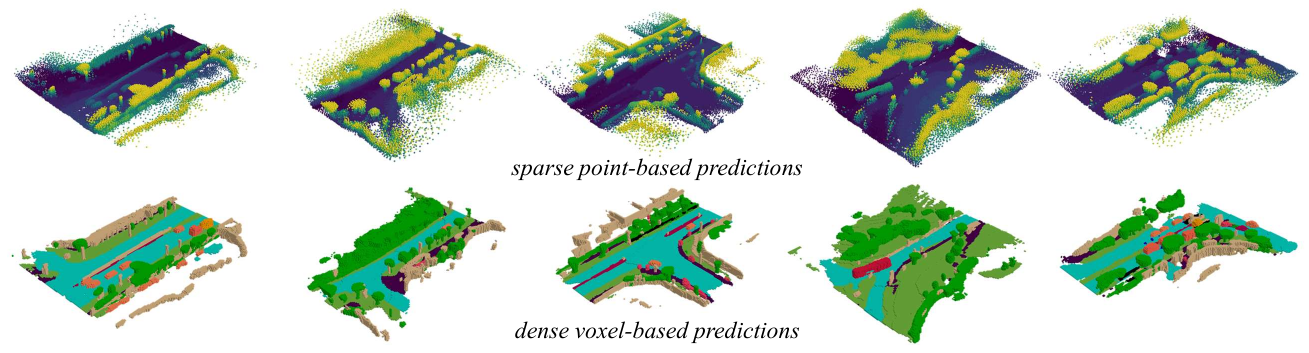}
    \caption{
        Visualizations of the sparse point-based and dense voxel-based predictions of \model{}.
    }\label{fig:vis1}
\end{figure*}

\begin{figure*}[t]
    \centering
    \includegraphics[width=\textwidth]{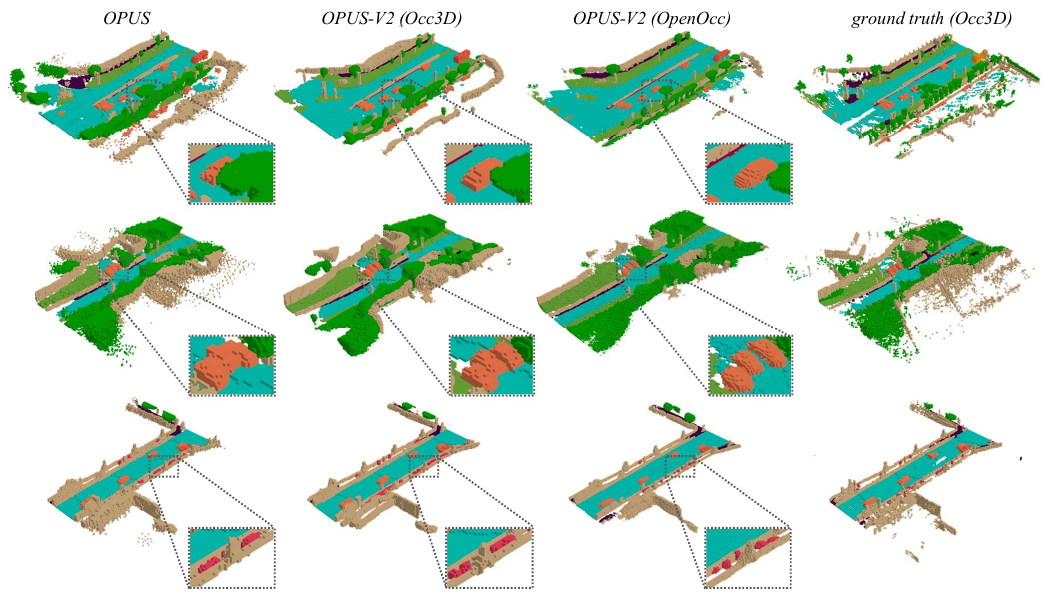}
    \caption{
        Results of the original OPUS and \model{} trained on Occ3D and OpenOccupancy.
    }\label{fig:vis2}
\end{figure*}

\subsection{Visualization}
In this section, we visualize results of \model{}.

\subsubsection{Sparse and dense predictions}
We visualize  sparse point-based and dense voxel-based predictions of \model{} in \cref{fig:vis1}.
For sparse predictions, we illustrate the positions of the set of points.
As  \cref{fig:vis1}(above) shows, the predicted points roughly outlined the geometry of the environment, attributed to the supervision of the chamfer distance loss.
This allowed the model to focus on occupied regions, thereby notably improving its accuracy and efficiency.
However, the flexibility of the points also led to uneven density and edge noise.
Those problems were alleviated by the subsequent \smodule{} module.
The dense predictions illustrated in  \cref{fig:vis1}(below) indicate that \smodule{} is able to suppress noisy points in the sparse predictions and thereby produce improved results.

\subsubsection{Comparison}
We visualize  predictions of OPUS and \model{} within a $\pm40$~m range. 
As  \cref{fig:vis2} shows, \model{} produced more accurate results, reducing surface holes and edge noise, demonstrating the effectiveness of the \smodule{}. 
Furthermore, we compare the results of \model{} trained on Occ3D and OpenOccupancy.
The results in the second and third columns of \cref{fig:vis2} show that the model trained on OpenOccupancy produced sharper and more precise occupancy predictions, indicating its ability to learn finer geometric details from higher-resolution ground truth.

\section{Conclusion}

In this paper, we have introduced \model{}, which further extends the point-based paradigm for occupancy prediction.
The core of \model{} is the \smodule{} module, a learnable module that bridges the representation gap by transforming sparse decoder predictions into dense voxel-based occupancy. 
This approach enables direct, end-to-end generation of the required dense outputs, eliminating manual post-processing and significantly boosting performance. 
Furthermore, by decoupling feature extraction in the decoders from occupancy generation in the \smodule{} module, \model{} gains the flexibility to readily adapt its output resolution. 
Extensive experiments on the Occ3D and OpenOccupancy benchmarks show that \model{} outperforms prior state-of-the-art methods in accuracy and efficiency, while demonstrating superior scalability.

Currently, \model{} also faces some new challenges.
The sparse convolutions in the \smodule{} module make edge device deployment more difficult.
In future , we plan to replace the sparse convolutions with 2D convolutions, as in FlashOcc, to facilitate deployment on edge devices.

% \section{Declarations}

% \subsection{Availability of Data and Materials}
% Our code is built on top of the codebase provided by SparseBEV, which is subject to the MIT license. Our experiments were conducted on  Occ3D and OpenOccupancy, which provide occupancy labels for nuScenes. Occ3D, OpenOccupancy, and nuScenes were released under the MIT, Apache-2.0, and CC BY-NC-SA 4.0 licenses, respectively.

% \subsection{Declaration of Competing Interests}
% Author Ming-Ming Cheng is on the editorial board of CVMJ.
% Other authors have no relevant financial or non-financial interests to disclose.

% \subsection{Funding}
% This work was supported by NSFC (62522607 and 62276145), and the Fundamental Research Funds for the Central Universities (Nankai University).

% \subsection{Author contributions}
% All authors contributed to the idea of this paper and the design of the model. 
% Material preparation, data collection, and analysis were performed by Jiabao Wang. 

% \subsection{Acknowledgements}
% We sincerely thank the editors and anonymous reviewers for their help in improving this paper.

% for bibtex
\bibliographystyle{CVMbib}
\bibliography{refs.bib}

\begin{thebibliography}{10}
\expandafter\ifx\csname urlstyle\endcsname\relax
  \providecommand{\doi}[1]{doi:\discretionary{}{}{}#1}\else
  \providecommand{\doi}{doi:\discretionary{}{}{}\begingroup
  \urlstyle{rm}\Url}\fi

\bibitem{tian2024occ3d}
Tian X, Jiang T, Yun L, Mao Y, Yang H, Wang Y, Wang Y, Zhao H. Occ3d: A
  large-scale 3d occupancy prediction benchmark for autonomous driving.
  \emph{Advances in Neural Information Processing Systems}, 2024, 36.

\bibitem{wang2023openoccupancy}
Wang X, Zhu Z, Xu W, Zhang Y, Wei Y, Chi X, Ye Y, Du D, Lu J, Wang X.
  Openoccupancy: A large scale benchmark for surrounding semantic occupancy
  perception. In \emph{Proceedings of the IEEE/CVF International Conference on
  Computer Vision}, 2023, 17850--17859.

\bibitem{cao2022monoscene}
Cao AQ, De~Charette R. Monoscene: Monocular 3d semantic scene completion. In
  \emph{Proceedings of the IEEE/CVF Conference on Computer Vision and Pattern
  Recognition}, 2022, 3991--4001.

\bibitem{li2023fb}
Li Z, Yu Z, Austin D, Fang M, Lan S, Kautz J, Alvarez JM. Fb-occ: 3d occupancy
  prediction based on forward-backward view transformation. \emph{arXiv
  preprint arXiv:2307.01492}, 2023.

\bibitem{zhang2023occformer}
Zhang Y, Zhu Z, Du D. Occformer: Dual-path transformer for vision-based 3d
  semantic occupancy prediction. In \emph{Proceedings of the IEEE/CVF
  International Conference on Computer Vision}, 2023, 9433--9443.

\bibitem{huang2023tri}
Huang Y, Zheng W, Zhang Y, Zhou J, Lu J. Tri-perspective view for vision-based
  3d semantic occupancy prediction. In \emph{Proceedings of the IEEE/CVF
  conference on computer vision and pattern recognition}, 2023, 9223--9232.

\bibitem{liu2023fully}
Liu H, Chen Y, Wang H, Yang Z, Li T, Zeng J, Chen L, Li H, Wang L. Fully sparse
  3d occupancy prediction. In \emph{European Conference on Computer Vision},
  2024, 54--71.

\bibitem{wang2024opus}
Wang J, Liu Z, Meng Q, Yan L, Wang K, Yang J, Liu W, Hou Q, Cheng MM. Opus:
  occupancy prediction using a sparse set. \emph{Advances in Neural Information
  Processing Systems}, 2024, 37: 119861--119885.

\bibitem{dang2025sparseworld}
Dang C, Liu H, Bao G, An P, Tang X, Ma J, Sun B, Wang Y. SparseWorld: A
  Flexible, Adaptive, and Efficient 4D Occupancy World Model Powered by Sparse
  and Dynamic Queries. \emph{arXiv preprint arXiv:2510.17482}, 2025.

\bibitem{li2025enhancing}
Li X, Zheng Y, Li P, Chen Y, Zhang YQ, Ding W. Enhancing Indoor Occupancy
  Prediction via Sparse Query-Based Multi-Level Consistent Knowledge
  Distillation. \emph{IEEE Robotics and Automation Letters}, 2025.

\bibitem{huang2024gau}
Huang Y, Zheng W, Zhang Y, Zhou J, Lu J. Gaussianformer: Scene as gaussians for
  vision-based 3d semantic occupancy prediction. In \emph{European Conference
  on Computer Vision}, 2024, 376--393.

\bibitem{huang2025gaussianformer}
Huang Y, Thammatadatrakoon A, Zheng W, Zhang Y, Du D, Lu J. Gaussianformer-2:
  Probabilistic gaussian superposition for efficient 3d occupancy prediction.
  In \emph{Proceedings of the Computer Vision and Pattern Recognition
  Conference}, 2025, 27477--27486.

\bibitem{kerbl3Dgaussians}
Kerbl B, Kopanas G, Leimk{\"u}hler T, Drettakis G. 3D Gaussian Splatting for
  Real-Time Radiance Field Rendering. \emph{ACM Transactions on Graphics},
  2023, 42(4).

\bibitem{barr1981superquadrics}
Barr AH. Superquadrics and angle-preserving transformations. \emph{IEEE
  Computer graphics and Applications}, 1981, 1(1): 11--23.

\bibitem{chen2025alocc}
Chen D, Fang J, Han W, Cheng X, Yin J, Xu C, Khan FS, Shen J. ALOcc: Adaptive
  Lifting-Based 3D Semantic Occupancy and Cost Volume-Based Flow Predictions.
  In \emph{Proceedings of the IEEE/CVF International Conference on Computer
  Vision}, 2025, 4156--4166.

\bibitem{elfes2002using}
Elfes A. Using occupancy grids for mobile robot perception and navigation.
  \emph{Computer}, 2002, 22(6): 46--57.

\bibitem{moravec1985high}
Moravec H, Elfes A. High resolution maps from wide angle sonar. In
  \emph{Proceedings. 1985 IEEE international conference on robotics and
  automation}, volume~2, 1985, 116--121.

\bibitem{zhou2018voxelnet}
Zhou Y, Tuzel O. Voxelnet: End-to-end learning for point cloud based 3d object
  detection. In \emph{Proceedings of the IEEE conference on computer vision and
  pattern recognition}, 2018, 4490--4499.

\bibitem{yan2018second}
Yan Y, Mao Y, Li B. Second: Sparsely embedded convolutional detection.
  \emph{Sensors}, 2018, 18(10): 3337.

\bibitem{graham2014spatially}
Graham B. Spatially-sparse convolutional neural networks. \emph{arXiv preprint
  arXiv:1409.6070}, 2014.

\bibitem{tesla}
Tesla. Tesla AI Day 2022. YouTube:
  \url{https://www.youtube.com/watch?v=ODSJsviD_SU}, 2022.

\bibitem{wei2023surroundocc}
Wei Y, Zhao L, Zheng W, Zhu Z, Zhou J, Lu J. Surroundocc: Multi-camera 3d
  occupancy prediction for autonomous driving. In \emph{Proceedings of the
  IEEE/CVF International Conference on Computer Vision}, 2023, 21729--21740.

\bibitem{tong2023scene}
Tong W, Sima C, Wang T, Chen L, Wu S, Deng H, Gu Y, Lu L, Luo P, Lin D, et~al..
  Scene as occupancy. In \emph{Proceedings of the IEEE/CVF International
  Conference on Computer Vision}, 2023, 8406--8415.

\bibitem{caesar2020nuscenes}
Caesar H, Bankiti V, Lang AH, Vora S, Liong VE, Xu Q, Krishnan A, Pan Y, Baldan
  G, Beijbom O. nuscenes: A multimodal dataset for autonomous driving. In
  \emph{Proceedings of the IEEE/CVF conference on computer vision and pattern
  recognition}, 2020, 11621--11631.

\bibitem{sun2020waymo}
Sun P, Kretzschmar H, Dotiwalla X, Chouard A, Patnaik V, Tsui P, Guo J, Zhou Y,
  Chai Y, Caine B, Vasudevan V, Han W, Ngiam J, Zhao H, Timofeev A, Ettinger S,
  Krivokon M, Gao A, Joshi A, Zhang Y, Shlens J, Chen Z, Anguelov D.
  Scalability in Perception for Autonomous Driving: Waymo Open Dataset. In
  \emph{Proceedings of the IEEE/CVF Conference on Computer Vision and Pattern
  Recognition (CVPR)}, 2020, 2446--2454.

\bibitem{behley2019semantickitti}
Behley J, Garbade M, Milioto A, Quenzel J, Behnke S, Stachniss C, Gall J.
  Semantickitti: A dataset for semantic scene understanding of lidar sequences.
  In \emph{Proceedings of the IEEE/CVF international conference on computer
  vision}, 2019, 9297--9307.

\bibitem{caesar2021nuplan}
Caesar H, Kabzan J, Tan KS, Fong WK, Wolff E, Lang A, Fletcher L, Beijbom O,
  Omari S. nuplan: A closed-loop ml-based planning benchmark for autonomous
  vehicles. \emph{arXiv preprint arXiv:2106.11810}, 2021.

\bibitem{yu2023flashocc}
Yu Z, Shu C, Deng J, Lu K, Liu Z, Yu J, Yang D, Li H, Chen Y. Flashocc: Fast
  and memory-efficient occupancy prediction via channel-to-height plugin.
  \emph{arXiv preprint arXiv:2311.12058}, 2023.

\bibitem{yu2024panoptic}
Yu Z, Shu C, Sun Q, Bian Y, Wei X, Yu J, Liu Z, Yang D, Li H, Chen Y.
  Panoptic-flashocc: An efficient baseline to marry semantic occupancy with
  panoptic via instance center. \emph{arXiv preprint arXiv:2406.10527}, 2024.

\bibitem{ma2023cotr}
Ma Q, Tan X, Qu Y, Ma L, Zhang Z, Xie Y. Cotr: Compact occupancy transformer
  for vision-based 3d occupancy prediction. In \emph{Proceedings of the
  IEEE/CVF Conference on Computer Vision and Pattern Recognition}, 2024,
  19936--19945.

\bibitem{li2023voxformer}
Li Y, Yu Z, Choy C, Xiao C, Alvarez JM, Fidler S, Feng C, Anandkumar A.
  Voxformer: Sparse voxel transformer for camera-based 3d semantic scene
  completion. In \emph{Proceedings of the IEEE/CVF conference on computer
  vision and pattern recognition}, 2023, 9087--9098.

\bibitem{zheng2024gaussianad}
Zheng W, Wu J, Zheng Y, Zuo S, Xie Z, Yang L, Pan Y, Hao Z, Jia P, Lang X,
  et~al.. Gaussianad: Gaussian-centric end-to-end autonomous driving.
  \emph{arXiv preprint arXiv:2412.10371}, 2024.

\bibitem{sun2024gsrender}
Sun Q, Shu C, Zhou S, Yu Z, Chen Y, Yang D, Chun Y. Gsrender: Deduplicated
  occupancy prediction via weakly supervised 3d gaussian splatting. \emph{arXiv
  preprint arXiv:2412.14579}, 2024.

\bibitem{shi2025odg}
Shi Y, Zhu Y, Han S, Jeong J, Ansari A, Cai H, Porikli F. ODG: Occupancy
  Prediction Using Dual Gaussians. \emph{arXiv preprint arXiv:2506.09417},
  2025.

\bibitem{liu2023sparsebev}
Liu H, Teng Y, Lu T, Wang H, Wang L. Sparsebev: High-performance sparse 3d
  object detection from multi-camera videos. In \emph{Proceedings of the
  IEEE/CVF International Conference on Computer Vision}, 2023, 18580--18590.

\bibitem{carion2020end}
Carion N, Massa F, Synnaeve G, Usunier N, Kirillov A, Zagoruyko S. End-to-end
  object detection with transformers. In \emph{European conference on computer
  vision}, 2020, 213--229.

\bibitem{zhu2020deformable}
Zhu X, Su W, Lu L, Li B, Wang X, Dai J. Deformable detr: Deformable
  transformers for end-to-end object detection. In \emph{International
  Conference on Learning Representations}, 2021, 1--xxx.

\bibitem{meng2021conditional}
Meng D, Chen X, Fan Z, Zeng G, Li H, Yuan Y, Sun L, Wang J. Conditional detr
  for fast training convergence. In \emph{Proceedings of the IEEE/CVF
  international conference on computer vision}, 2021, 3651--3660.

\bibitem{liu2022dab}
Liu S, Li F, Zhang H, Yang X, Qi X, Su H, Zhu J, Zhang L. {DAB}-{DETR}: Dynamic
  Anchor Boxes are Better Queries for {DETR}. In \emph{International Conference
  on Learning Representations}, 2022, 1--xxx.

\bibitem{li2022dn}
Li F, Zhang H, Liu S, Guo J, Ni LM, Zhang L. Dn-detr: Accelerate detr training
  by introducing query denoising. In \emph{Proceedings of the IEEE/CVF
  conference on computer vision and pattern recognition}, 2022, 13619--13627.

\bibitem{wang2022anchor}
Wang Y, Zhang X, Yang T, Sun J. Anchor detr: Query design for transformer-based
  detector. In \emph{Proceedings of the AAAI conference on artificial
  intelligence}, volume~36, 2022, 2567--2575.

\bibitem{zhang2022dino}
Zhang H, Li F, Liu S, Zhang L, Su H, Zhu J, Ni LM, Shum HY. Dino: Detr with
  improved denoising anchor boxes for end-to-end object detection. In
  \emph{International Conference on Learning Representations}, 2023, 1--xxx.

\bibitem{sun2021rethinking}
Sun Z, Cao S, Yang Y, Kitani KM. Rethinking transformer-based set prediction
  for object detection. In \emph{Proceedings of the IEEE/CVF international
  conference on computer vision}, 2021, 3611--3620.

\bibitem{wang2022detr3d}
Wang Y, Guizilini VC, Zhang T, Wang Y, Zhao H, Solomon J. Detr3d: 3d object
  detection from multi-view images via 3d-to-2d queries. In \emph{Conference on
  Robot Learning}, 2022, 180--191.

\bibitem{liu2022petr}
Liu Y, Wang T, Zhang X, Sun J. Petr: Position embedding transformation for
  multi-view 3d object detection. In \emph{European Conference on Computer
  Vision}, 2022, 531--548.

\bibitem{lin2022sparse4d}
Lin X, Lin T, Pei Z, Huang L, Su Z. Sparse4d: Multi-view 3d object detection
  with sparse spatial-temporal fusion. \emph{arXiv preprint arXiv:2211.10581},
  2022.

\bibitem{wang2023exploring}
Wang S, Liu Y, Wang T, Li Y, Zhang X. Exploring object-centric temporal
  modeling for efficient multi-view 3d object detection. In \emph{Proceedings
  of the IEEE/CVF International Conference on Computer Vision}, 2023,
  3621--3631.

\bibitem{tang2024sparseocc}
Tang P, Wang Z, Wang G, Zheng J, Ren X, Feng B, Ma C. Sparseocc: Rethinking
  sparse latent representation for vision-based semantic occupancy prediction.
  In \emph{Proceedings of the IEEE/CVF Conference on Computer Vision and
  Pattern Recognition}, 2024, 15035--15044.

\bibitem{2019arXiv190809492Z}
{Zhu} B, {Jiang} Z, {Zhou} X, {Li} Z, {Yu} G. {Class-balanced Grouping and
  Sampling for Point Cloud 3D Object Detection}. \emph{arXiv e-prints}, 2019:
  arXiv:1908.09492.

\bibitem{pan2023renderocc}
Pan M, Liu J, Zhang R, Huang P, Li X, Xie H, Wang B, Liu L, Zhang S. Renderocc:
  Vision-centric 3d occupancy prediction with 2d rendering supervision. In
  \emph{2024 IEEE International Conference on Robotics and Automation (ICRA)},
  2024, 12404--12411.

\bibitem{li2022bevformer}
Li Z, Wang W, Li H, Xie E, Sima C, Lu T, Qiao Y, Dai J. Bevformer: Learning
  bird’s-eye-view representation from multi-camera images via spatiotemporal
  transformers. In \emph{European conference on computer vision}, 2022, 1--18.

\bibitem{huang2021bevdet}
Huang J, Huang G, Zhu Z, Ye Y, Du D. Bevdet: High-performance multi-camera 3d
  object detection in bird-eye-view. \emph{arXiv preprint arXiv:2112.11790},
  2021.

\bibitem{liao2025stcocc}
Liao Z, Wei P, Chen S, Wang H, Ren Z. Stcocc: Sparse spatial-temporal cascade
  renovation for 3d occupancy and scene flow prediction. In \emph{Proceedings
  of the Computer Vision and Pattern Recognition Conference}, 2025, 1516--1526.

\bibitem{yang2024daocc}
Yang Z, Dong Y, Wang J, Wang H, Ma L, Cui Z, Liu Q, Pei H, Zhang K, Zhang C.
  DAOcc: 3D Object Detection Assisted Multi-Sensor Fusion for 3D Occupancy
  Prediction. \emph{IEEE Transactions on Circuits and Systems for Video
  Technology}, 2024, 36: 1742--1753.

\bibitem{duan2025sdgocc}
Duan Z, Dang C, Hu X, An P, Ding J, Zhan J, Xu Y, Ma J. SDGOCC: Semantic and
  Depth-Guided Bird's-Eye View Transformation for 3D Multimodal Occupancy
  Prediction. In \emph{Proceedings of the Computer Vision and Pattern
  Recognition Conference}, 2025, 6751--6760.

\bibitem{chen20203d}
Chen X, Lin KY, Qian C, Zeng G, Li H. 3d sketch-aware semantic scene completion
  via semi-supervised structure prior. In \emph{Proceedings of the IEEE/CVF
  Conference on Computer Vision and Pattern Recognition}, 2020, 4193--4202.

\bibitem{li2020anisotropic}
Li J, Han K, Wang P, Liu Y, Yuan X. Anisotropic convolutional networks for 3d
  semantic scene completion. In \emph{Proceedings of the IEEE/CVF Conference on
  Computer Vision and Pattern Recognition}, 2020, 3351--3359.

\bibitem{roldao2020lmscnet}
Roldao L, De~Charette R, Verroust-Blondet A. Lmscnet: Lightweight multiscale 3d
  semantic completion. In \emph{2020 International Conference on 3D Vision
  (3DV)}, 2020, 111--119.

\bibitem{yan2021sparse}
Yan X, Gao J, Li J, Zhang R, Li Z, Huang R, Cui S. Sparse single sweep lidar
  point cloud segmentation via learning contextual shape priors from scene
  completion. In \emph{Proceedings of the AAAI conference on artificial
  intelligence}, 2021, 3101--3109.

\bibitem{he2016deep}
He K, Zhang X, Ren S, Sun J. Deep residual learning for image recognition. In
  \emph{Proceedings of the IEEE conference on computer vision and pattern
  recognition}, 2016, 770--778.

\bibitem{loshchilov2018decoupled}
Loshchilov I, Hutter F. Decoupled Weight Decay Regularization. In
  \emph{International Conference on Learning Representations}, 2018, 1--xxx.

\bibitem{loshchilov2016sgdr}
Loshchilov I, Hutter F. SGDR: STOCHASTIC GRADIENT DESCENT WITH WARM RESTARTS.
  In \emph{International Conference on Learning Representations}, 2017, 1--xxx.

\end{thebibliography}

\end{document}